\documentclass[10pt,twocolumn,letterpaper]{article}

\usepackage{cvpr}
\usepackage{times}
\usepackage{epsfig}
\usepackage{graphicx}
\usepackage{amsmath}
\usepackage{amssymb}
\usepackage{multirow}
\usepackage{color}
\usepackage{diagbox}

\usepackage[breaklinks=true,bookmarks=false]{hyperref}
\cvprfinalcopy 
\def\cvprPaperID{7199} 

\ifcvprfinal\fi
\begin{document}

\title{Universal Weighting Metric Learning for Cross-Modal Matching}

\author{Jiwei Wei, Xing Xu\thanks{Corresponding author. }, Yang Yang, Yanli Ji, Zheng Wang, Heng Tao Shen\\
Center for Future Media \& School of Computer Science and Engineering,\\
University of Electronic Science and Technology of China, China\\
}
\maketitle

\begin{abstract}
Cross-modal matching has been a highlighted research topic in both vision and language areas. Learning appropriate mining strategy to sample and weight informative pairs is crucial for the cross-modal matching performance. However, most existing metric learning methods are developed for unimodal matching, which is unsuitable for cross-modal matching on multimodal data with heterogeneous features. To address this problem, we propose a simple and interpretable universal weighting framework for cross-modal matching, which provides a tool to analyze the interpretability of various loss functions. Furthermore, we introduce a new polynomial loss under the universal weighting framework, which defines a weight function for the positive and negative informative pairs respectively. Experimental results on two image-text matching benchmarks and two video-text matching benchmarks validate the efficacy of the proposed method.  The source code is available at:\textcolor[rgb]{1,0,0}{ https://github.com/wayne980/PolyLoss}. 

\end{abstract}

\section{Introduction}
Cross-modal matching aims at retrieving relevant instances of a different media type from the query, which has a variety of applications such as Image-Text matching~\cite{faghrivse++, wang2019matching, shen2020exploiting, xu2019ternary, wang2017adversarial, xu2017learning, zhang2019optimal, cui2019scalable, hu2018collective}, Video-Text matching~\cite{ Liu2019a, song2018deterministic, gao2019hierarchical, yang2018video}, Sketch-based image matching~\cite{dey2019doodle}, etc. Compared with unimodal matching, cross-modal matching is more challenging due to the heterogeneous gap between different modalities. The key issue in cross-modal matching is to reduce the heterogeneous gap and exploit discriminative information across modalities~\cite{liong2016deep,gan2016learning,xu2019deep,wei2019residual, shen2019scalable}.
\begin{figure}[t]
\includegraphics[width=0.9\linewidth]{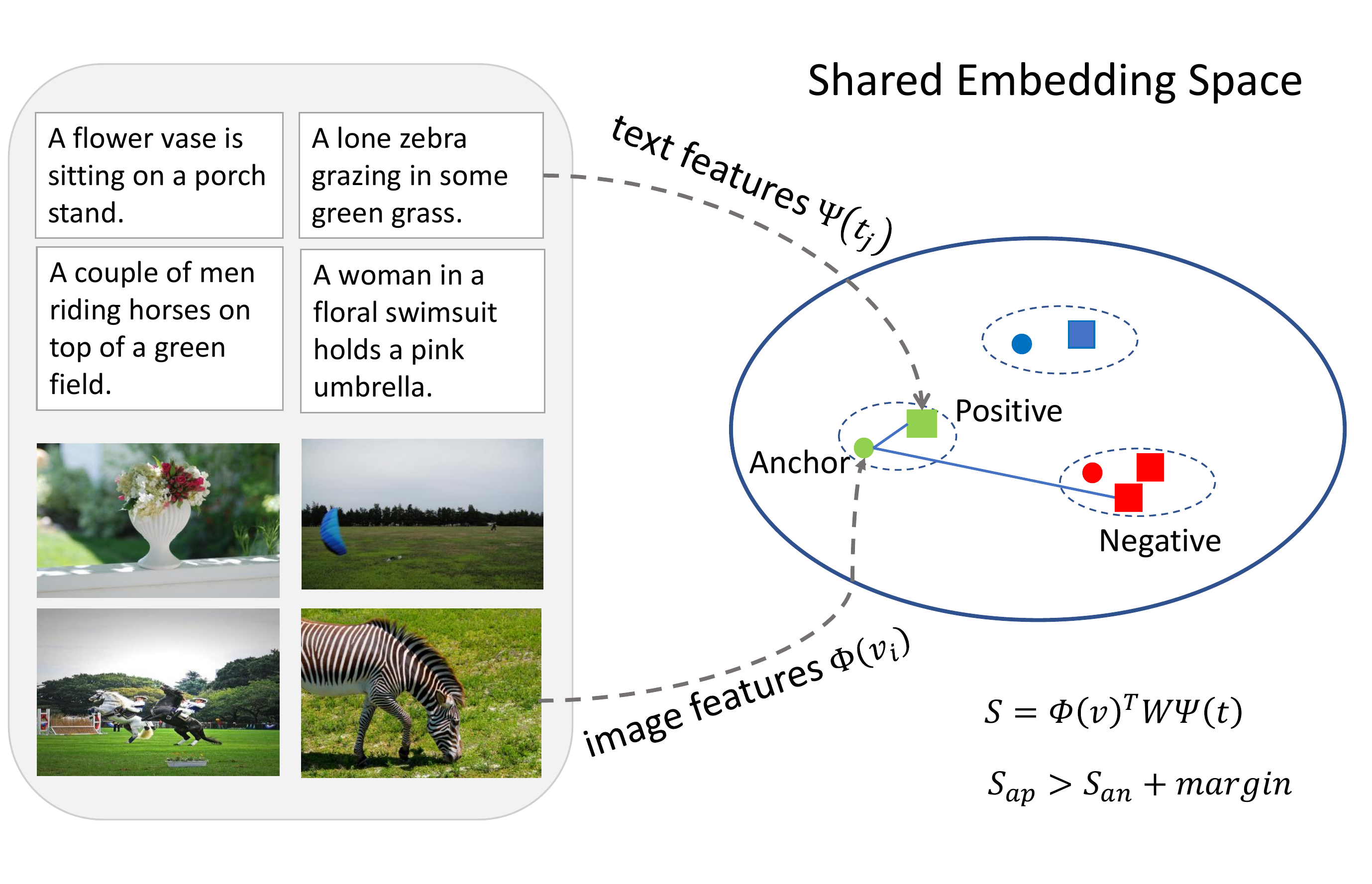}
\caption{A typical solution for cross-modal matching is to learn a shared embedding space where visual features $\Phi (v)$ and text features $\Psi(t)$ can be compared. Points with the same shape are from the same modality. A triplet loss was utilized to encourage the similarity of positive (matching) pairs larger than negative (non-matching) pairs. Take image-text matching as an example.}
\label{fig1}
\end{figure}

A common solution for cross-modal matching is to learn a shared embedding space for different modalities so that the features from different modalities can be compared. Recently, a variety of cross-modal matching methods have been devoted to learning richer semantic representations for different modalities and a ranking loss is adopted to jointly optimize the network so that the similarity of the positive pairs is higher than that of all negative pairs, as illustrated in Figure~\ref{fig1}. In previous literature, attention mechanism \cite{lee2018stacked} and generative models \cite{gu2018look,huang2018learning} have been explored to build advanced encoding networks.
Liu \etal \cite{liu2017learning} proposed a recurrent residual fusion block to reduce the modality gap, and a triplet loss \cite{hoffer2015deep} was used to encourage semantically associated samples close to each other in the shared embedding space. Li \etal \cite{li2019visual} proposed a visual reasoning model to generate global representation of a scene.

\begin{figure}
\includegraphics[width=\linewidth]{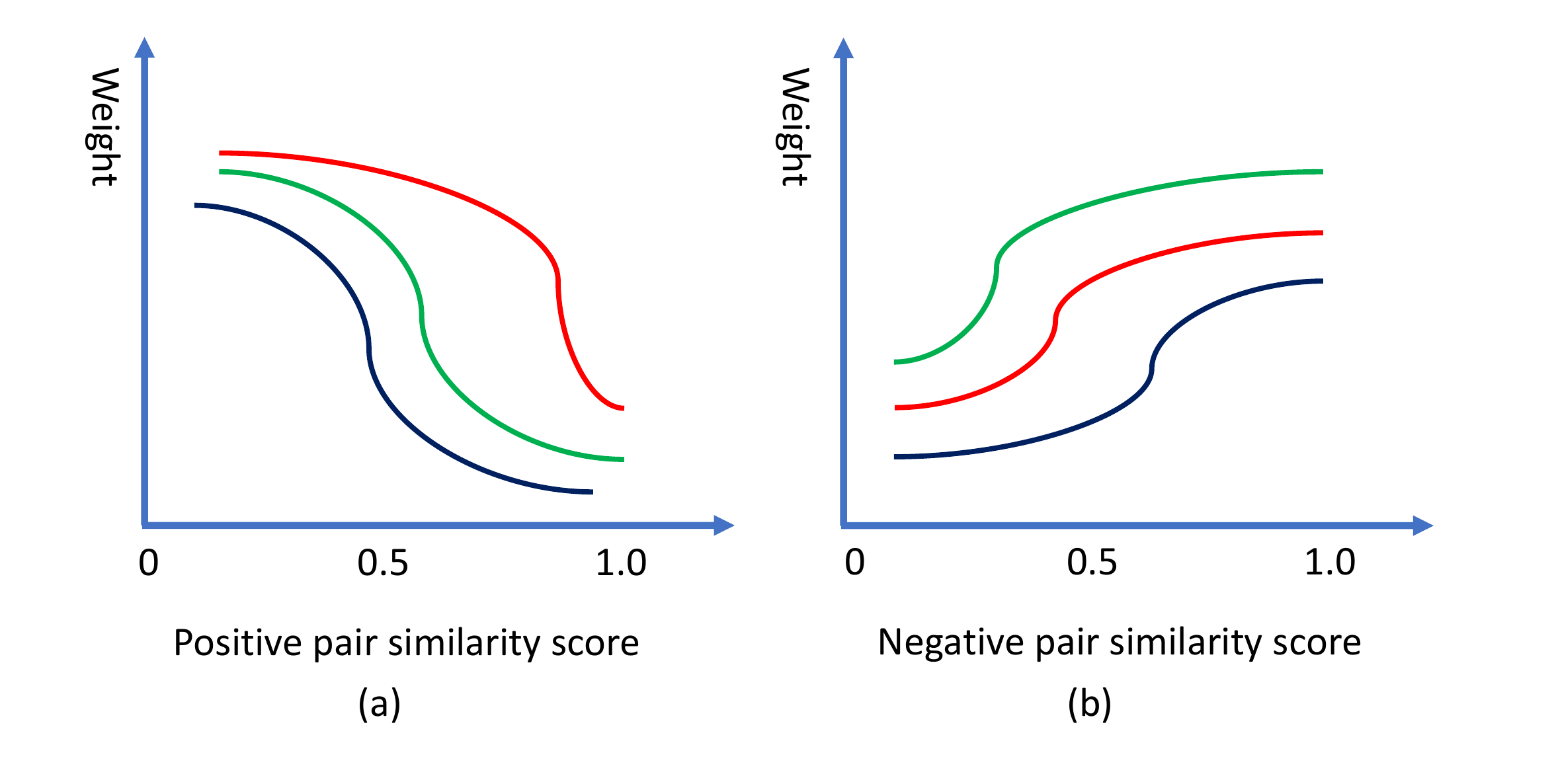}
\caption{As the positive pairs similarity score increases, its weight value decreases; As the negative pairs similarity score increases, its weight value increases. }
\label{fig2}
\end{figure}

While these methods have achieved encouraging performance, most of them use the ranking loss as an objective function, which usually trained with random sampling. This gives rise to an issue for cross-modal matching, where random sampling cannot effectively select informative pairs for training, leading a slow convergence and poor performance. While more recent metric learning methods have provided various mining strategies for unimodal matching, few of them are suitable for cross-modal matching. Hence, learning an appropriate mining strategy to sample and weight informative pairs is still a challenging problem for cross-modal matching.
In this paper, we propose a universal weighting framework for cross-modal matching. Our intuition is based on the fact that a larger weight is assigned to a more informative pair, as illustrated in Figure~\ref{fig2}. Unlike widely used unweighted triplet loss which treats all pairs equally, our proposed universal weighting framework can effectively assign appropriate weight to informative pairs for cross-modal matching. Specifically, we define two polynomial functions to calculate the weight values for positive and negative pairs respectively. Furthermore, we introduce a new polynomial loss under the universal weighting framework. Since the form of a polynomial function is flexible, our polynomial loss has a better generalization.


The major contributions of this paper are summarized as follows:


\begin{itemize}
\item We propose a universal weighting framework for cross-modal matching, which defines two polynomial functions to calculate the weight values for positive and negative pairs respectively. It provides a powerful tool to analyze the interpretability of various loss functions.
\item We introduce a new polynomial loss under the universal weighting framework. The polynomial loss can effectively select informative pairs from redundant pairs, and assign appropriate weights to different pairs, resulting in performance boost.

\item We conduct extensive experiments and evaluate our proposed method on two cross-modal matching tasks, image-text matching and video-text matching. Experimental results demonstrate that our method achieves very competitive performance on the four widely used benchmark datasets: MS-COCO, Flickr30K, ActivityNet-captions and MSR-VTT.
\end{itemize}

\section{Related Work}
{\bfseries Cross-Modal Matching.} Cross-modal matching has a variety of applications, such as Image-Text matching~\cite{faghrivse++,wang2019matching}, Video-Text matching~\cite{gan2016recognizing,song2019polysemous,Liu2019a}, Sketch-based image retrieval~\cite{dey2019doodle} etc. The key issue of cross-modal matching is measuring the similarity between different modal features. A common solution is to learn a shared embedding space where features of different modalities can be directly compared. In recent years, a variety of methods have been devoted to learning modality invariant features.


Lee \etal~\cite{lee2018stacked} proposed a stacked cross attention network for image-text matching, which measures the image-text similarity by aligning image regions and words. Li \etal~\cite{li2019visual} used graph convolutional network to generate relationship-enhanced image region features, then a global semantic reasoning network is performed to generate discriminative visual features that capture key objects and semantic concepts of a scene. Song \etal\cite{song2019polysemous} introduced a polysemous instance embedding network that uses multi-head self-attention and residual learning to generate multiple representations of an instance. Liu \etal~\cite{Liu2019a} proposed a collaborative expert (CE) framework for video-text matching, which generates dense representations for videos via aggregating information from different pre-trained models. Above embedding-based methods learn an advanced encoding network to generate richer semantic representations for different modalities, which make the matched pairs close to each other and the mismatched pairs far apart in the shared embedding space.

{\bfseries Metric Learning for Cross-Modal Matching.} Another popular approach for cross-modal matching is to learn a loss function in the embedding space, which encourages the similarity of matched pairs larger than mismatched pairs. In recent years, a variety of metric learning methods have been proposed in both vision and language areas. However, most of existing metric learning methods are designed for unimodal matching, which cannot effectively model the relationship of features captured from different modalities~\cite{liong2016deep}. Only few of metric learning methods have been implemented particularly for cross-modal matching~\cite{xu2019deep, liong2016deep, faghrivse++}.

Liong \etal~\cite{liong2016deep} introduced a deep coupled metric learning that designs two nonlinear transformations to reduce the modality map. Frome \etal~\cite{frome2013devise} proposed a deep visual-semantic embedding model mapping visual features and semantic features into a shared embedding space, using a hinge rank loss as the objective function. Faghri \etal~\cite{faghrivse++} introduced a variant triplet loss for image-text matching, and reported improved results. Xu \etal \cite{xu2019deep} introduced a modality classifier to ensure that the transformed features are statistically indistinguishable. However, these methods treat positive and negative pairs equally. Hardly any advanced sampling and weighting mechanism has been proposed for cross-modal matching. In this work, we present a universal weighting framework for cross-modal matching, which assigns a larger weight value to a harder sample.

\section{The Proposed Approach}

In this section, we formulate the sampling problem of cross-modal matching as a general weighting formulation. The proposed polynomial loss will be elaborated afterward.
\subsection{Problem Statement}
Let $v_i \in \mathbb{R}^{d_1}$ be a visual feature vector, $t_i \in \mathbb {R}^{d_2}$ be a text feature vector, $D=\{(v_i, t_i)\}$ be a training set of cross-modal instance pairs. In general, components of an instance pair come from different modalities. For simplicity, we refer to $(v_i, t_i)$ as a positive pair and $(v_i, t_{j, i\neq j})$ as a negative pair. Given a query instance, the goal of cross-modal matching is to find a sample that matches it in another modality gallery. In the case of image-text matching, given an image caption $t_i$, the goal is to find the most relevant image $v_i$ in the image gallery. It is important to note that in the cross-modal matching task, there is only one positive sample for each anchor in a mini-batch. Previous work for cross-modal matching focused on building a shared embedding space that contains both the image and text. The core idea behind these methods is that there exists a mapping function, $S(v,t; W)=\Phi (v)^TW\Psi (t)$ to measure the similarity score between the visual features $\Phi (v)$ and the text features $\Psi (t)$. $W$ is the parameter of $S$. In general, the similarity score of the positive pair is higher than the negative pair by a margin, it can be formulated as:
\begin{equation}
S(v_i,t_i)>S(v_i,t_{j, j\neq i})+\lambda_0, \forall  v_i,
\label{eq1}
\end{equation}
\begin{equation}
S(v_j,t_j)>S(v_{i, i\neq j},t_j)+\lambda_0 , \forall  t_j,
\label{eq2}
\end{equation}
where $\lambda_0$ is a fixed margin.

\begin{figure*}
\includegraphics[width=\linewidth]{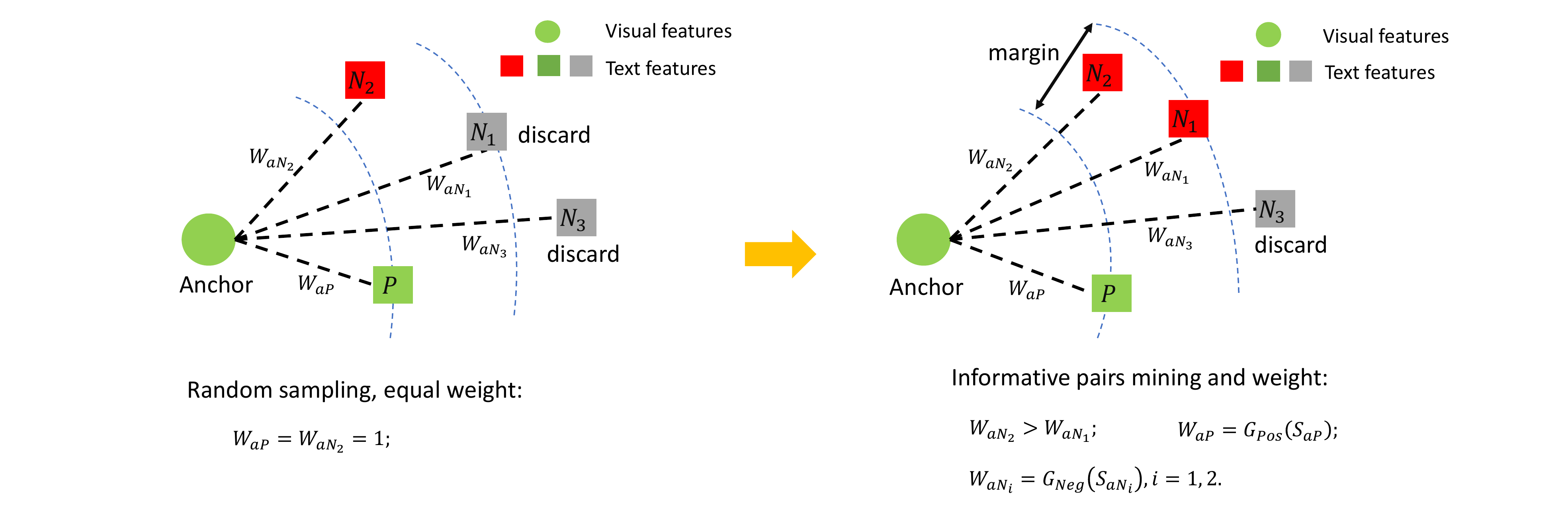}
\caption{Illustration of our informative pairs mining and universal weighting framework for cross-modal matching. Points with the same shape are from the same modality. $P$ is the only positive sample of anchor, $N_1$, $N_2$ and $N_3$ are the negative samples of anchor. Left: An example of random sampling and equal weighting; Right: The proposed negative pairs mining and universal weighting framework for cross-modal matching;}
\label{fig3}
\end{figure*}

Since cross-modal matching is a mutual retrieval problem, the widely used triplet loss is formulated as :
\begin{equation}
L=[S(v,\hat{t})-S(v,t)+\lambda_0]_++[S(\hat{v},t)-S(v,t)+\lambda_0]_+,
\label{eq3}
\end{equation}
where $(v,t)$ is positive pair, $(v,\hat{t}$) is the hardest negative pair for a query $v$, and  $(\hat{v},t)$ is the hardest negative pair for a query caption $t$. $[x]_+ = max(x,0)$. However, these methods discard pairs with less information than the hardest pair, while treating positive pairs and negative pairs equally. To our best knowledge, there is no advanced sampling and weighting method for cross-modal matching.

\subsection{Universal Weighting Framework for Cross-Modal Matching}

Let $N_{v_i}=\{S_{ij,i\neq j}\}$ be the set of similarity scores for all negative pairs of a sample $v_i$, and $N_{t_j}=\{S_{ij,j\neq i}\}$ be the set of similarity scores for all negative pairs of a sample $t_j$. Most existing hinge-based loss functions $L$ can be formulated as a function of similarity scores: $L(\{S_{ij}\})$. Current existing weighting methods are given a special function to represent the relationship between weight values and similarity scores, the form of the function varies from task to task. All these functions can be reformulated into a universal weighting framework:

\begin{equation}
L=\sum_{i=1}^{i=N}{\{G_{Pos}S_{ii}+\sum{(G_{Neg}S_{ij, i\neq j})}\}},
\label{eq4}
\end{equation}
where $G_{Pos}$ is the weight value of the positive pair, $G_{Neg}$ is the weight value of negative pairs. Both of $G_{Pos}$ and $G_{Neg}$ are a function of similarity scores, but in a different forms.
\begin{equation}
G_{Pos}=G(S_{ii},N_{v_i}),
\label{eq5}
\end{equation}
\begin{equation}
G_{Neg}=G(S_{jj},N_{t_j}),
\label{eq6}
\end{equation}
where $G(\cdot )$ is a function that represents the relationship between weight value and similarity score. Theoretically, $G(\cdot )$ can be a function of self-similarity and relative similarity. The form of $G(\cdot)$ is various, but it should satisfy a basic rule: as the positive pairs' similarity score increases, its weight value decreases, and as the negative pairs' similarity score increases, its weight value increases. As illustrated in Figure \ref{fig2}. It provides a powerful tool to analyze the interpretability of various loss functions through weight analysis. Eq. \ref{eq4} is a general pair formulation, existing pair-based loss is one of its special cases.

\subsection{Informative Pairs Mining}
For cross-modal matching tasks, in a mini-batch, each anchor has only one positive sample, but many negative pairs. These negative pairs are redundant and most of them are less informative. Random sampling are difficult to select more informative pairs, resulting in the model is difficult to convergence and have a poor performance. It is urgent and important to develop efficient algorithms that can select informative negative pairs and discard less informative negative pairs. In this section, we select informative negative pairs by comparing the relative similarity scores between positive and negative pairs of an anchor.
For a given anchor $v_i$, we assume its positive sample is $t_i$, negative samples is $t_{j,i\neq j}$, and a negative pair $(v_i,t_j)$ is selected if $S_{ij}$ satisfies the condition:
\begin{equation}
S_{ij,i\neq j}>S_{ii}-\lambda ,
\label{eq7}
\end{equation}
where $\lambda$ is a fixed margin. As illustrated in Figure \ref{fig3}. Note that there only one positive sample for each anchor in a mini-batch.

\subsection{Polynomial Loss for Cross-Modal Matching}
Through the above steps, negative pairs with more informative pairs can be selected and less informative pairs can be discarded. In this section, we introduce a new weighting function to weight the selected pairs. Theoretically, $G(\cdot )$ can be a function of self-similarity and relative similarity. However, the more complex the $G(\cdot )$, the more hyper-parameters it contains, and the hyper-parameters setting is more difficult. In this paper, to reduce the number of hyper-parameters, we define $G(\cdot )$ as a function of self-similarity. Specifically, given a selected positive pair $(v_i, t_i)$, its weight $G_{Pos}$ can be formulated as:
\begin{equation}
G_{Pos}=a_mS_{ii}^{m}+a_{m-1}S_{ii}^{m-1}+\cdots+a_1S_{ii}+a_0,
\label{eq8}
\end{equation}
where $S_{ii}$ is the similarity score, $\{a_i\}_{i=0}^{i=m}$ is hyper-parameters, $m$ is positive integer. The form of $G_{Pos}$ is diverse, but its value should decrease with the increase of similarity score $S_{ii}$. Its trend should conform to the curve in figure \ref{fig2}a.

The weight $G_{Neg}$ for a selected negative pair $(v_i, t_j)$ can be formulated as:
\begin{equation}
G_{Neg}=b_kS_{ij}^{k}+b_{k-1}S_{ij}^{k-1}+\cdots+b_1S_{ij}+b_0,  i\neq j,
\label{eq9}
\end{equation}
where $S_{ij}$ is similarity score, $\{b_i\}_{i=0}^{i=k}$ is hyper-parameters, and $k$ is positive integer. The form of $G_{Neg}$ is diverse, but its trend should conform to the curve in figure \ref{fig2}b.

Through Eq. \ref{eq8} and \ref{eq9}, we obtain the weights of positive and negative pairs. In this paper, we introduce two different functions, average polynomial loss and maximum polynomial loss.

{\bfseries Avg Polynomial Loss.} Average polynomial loss can be defined as:
\begin{equation}
\begin{split}
L_{Avg}=\frac{1}{N}\sum_{i=1}^{i=N}[G_{Pos}S_{ii}^p+\frac{\sum_{S_{ij}\in N_{v_i}}{G_{Neg}S_{ij}^q}}{Num(N_{v_i})}+\lambda_1]_++\\
\frac{1}{N}\sum_{j=1}^{j=N}[G_{Pos}S_{jj}^p+\frac{\sum_{S_{ij}\in N_{t_j}}{G_{Neg}S_{ij}^q}}{Num(N_{t_j})}+\lambda_2]_+,\\
\end{split}
\label{eq10}
\end{equation}
Eq. \ref{eq10} can be reformulated as:
\begin{equation}
\begin{split}
L_{Avg}=\frac{1}{N}\sum_{i=1}^{i=N}[\sum^{P}{a_pS_{ii}^{p}}+\frac{\sum_{S_{ij}\in N_{v_i}}{\sum^{Q}{b_qS_{ij}^q}}}{Num(N_{v_i})}]_++\\
\frac{1}{N}\sum_{j=1}^{j=N}[\sum^{P}{a_pS_{jj}^{p}}+\frac{\sum_{S_{ij}\in N_{t_j}}{\sum^{Q}{b_qS_{ij}^q}}}{Num(N_{t_j})}]_+,\\
\end{split}
\label{eq11}
\end{equation}
here, $Num(N_{v_i})$ and $Num(N_{t_j})$ denote the number of negative pairs of sample $v_i$ and $t_j$ respectively. $P$ and $Q$ are the highest power of positive and negative pairs, respectively. Note, we make the minimum of $p$ and $q$ to 0, and $a_0=\lambda_1$, $b_0=\lambda_2$.

For cross-modal matching tasks, only one positive sample for each anchor in a mini-batch. Our loss function can make full use of informative negative pairs. Since cross-modal matching tasks involve the mutual retrieval between different modalities, our loss comprises two terms. The former term represents the loss of image retrieval caption, and the latter represents the loss of caption retrieval image. Since the form of a polynomial function is flexible, our polynomial loss has a better generalization.


{\bfseries Max Polynomial Loss.} To further highlight the superiority of our weighting mechanism, we introduce another version of polynomial loss $L_{Max}$, which only contains the hardest negative pair. The formulation of $L_{Max}$ is defined as:
\begin{equation}
\begin{split}
L_{Max}=\frac{1}{N}\sum_{i=1}^{i=N}[G_{Pos}S_{ii}^p+G_{Neg}Max\{N_{v_i}\}^q+\lambda_1]_++\\
\frac{1}{N}\sum_{j=1}^{j=N}[G_{Pos}S_{jj}^p+G_{Neg}Max\{N_{t_j}\}^q+\lambda_2]_+,\\
\end{split}
\label{eq12}
\end{equation}
here, $Max\{N_{v_i}\}$ and $Max\{N_{t_j}\}$ represent the hardest negative pair of sample $v_i$ and $t_j$ respectively. Eq.\ref{eq12} can be reformulated as:
\begin{equation}
\begin{split}
L_{Max}=\frac{1}{N}\sum_{i=1}^{i=N}[\sum^{P}{a_pS_{ii}^p}+\sum^{Q}{b_qMax\{N_{v_i}\}^q}]_++\\
\frac{1}{N}\sum_{j=1}^{j=N}[\sum^{P}{a_pS_{jj}^p}+\sum^{Q}{b_qMax\{N_{t_j}\}^q}]_+,\\
\end{split}
\label{eq13}
\end{equation}
Both $L_{Avg}$ and $L_{Max}$ can be minimized with gradient descent optimization. More discussions about $L_{Avg}$ and $L_{Max}$ can be found in the subsection of experiments. 

\section{Experiments}
In this section, we conducted extensive experiments to evaluate the proposed polynomial loss in both image-text matching and video-text matching tasks. Following the \cite{song2019polysemous, lee2018stacked}, we use the Recall@K as the performance metrics for both image-text matching and video-text matching tasks, which indicates the percentage of queries for which the model returns the correct item in its top K results. Ablation studies are conducted to analyze the effectiveness of proposed polynomial loss. We set the margin $\lambda$ in Eq. \ref{eq7} to 0.2 for all experiments.


\begin{table*}
\begin{center}
\setlength{\tabcolsep}{3mm}{
\begin{tabular}{l|l|ccc|ccc}
\hline
\multirow{2}{*}{Methods} & \multirow{2}{*}{Loss Function} &\multicolumn{3}{c}{Image-to-Text} & \multicolumn{3}{c}{ Text-to-Image }\\
\cline{3-8}
&&R@1&R@5&R@10&R@1 &R@5&R@10\\
\hline

RRF \cite{liu2017learning}& Triplet &47.6&77.4&87.1&35.4&68.3&79.9\\
VSE++~\cite{faghrivse++}& Triplet &52.9 &80.5& 87.2& 39.6& 70.1 &79.5\\
DAN \cite{nam2017dual}& Triplet &55.0& 81.8& 89.0& 39.4& 69.2& 79.1\\
SCO~\cite{huang2018learning}&Triplet+NLL &55.5&82.0&89.3& 41.1&70.5&80.1\\
SCAN (I2T)~\cite{lee2018stacked}& Triplet&67.9&89.0&94.4&43.9&74.2&82.8 \\
{\bfseries SCAN (I2T)}&{\bfseries Max Polynomial Loss}&{\bfseries 69.4}&{\bfseries 89.9}&{\bfseries 95.4}&{\bfseries 47.5}&{\bfseries 75.5}&{\bfseries 83.1}\\
\hline
\end{tabular}}
\end{center}

\caption{Experimental results on Flickr30K.}
\label{t1}
\end{table*}

\begin{table*}

\begin{center}
\setlength{\tabcolsep}{3mm}{
\begin{tabular}{l|l|ccc|ccc}
\hline
\multirow{2}{*}{Methods} & \multirow{2}{*}{Loss Function} &\multicolumn{3}{c}{Image-to-Text} & \multicolumn{3}{c}{ Text-to-Image }\\
\cline{3-8}
&&R@1&R@5&R@10&R@1 &R@5&R@10\\
\hline
\multicolumn{8}{c}{1K Test images}\\
\hline
VSE++~\cite{faghrivse++} & Triplet &64.6 &89.1& 95.7& 52.0& 83.1 &92.0\\
GXN \cite{gu2018look}& Triplet & 68.5 &- &97.9 &56.6 &-&{\bfseries 94.5}\\
PVSE~\cite{song2019polysemous}&Triplet+$L_{div}$+$L_{mmd}$&69.2 &91.6& 96.6& 55.2& 86.5& 93.7\\
SCAN (I2T)~\cite{lee2018stacked}& Triplet& 69.2&93.2&97.5&54.4&86.0&93.6\\
{\bfseries SCAN (I2T)} & {\bfseries Max Polynomial Loss}&{\bfseries 71.1}&{\bfseries 93.7} &{\bfseries 98.2}&{\bfseries 56.8}&{\bfseries 86.7}&93.0\\
\hline
\multicolumn{8}{c}{5K Test images}\\
\hline
VSE++~\cite{faghrivse++}&Triplet&41.3&71.1&81.2&30.3&59.4&72.4\\
GXN \cite{gu2018look}& Triplet & 42.0&- &84.7 &31.7 &-&74.6\\
PVSE~\cite{song2019polysemous}&Triplet+$L_{div}$+$L_{mmd}$&45.2&74.3&84.5&32.4&63.0&75.0\\
SCAN (I2T) \cite{lee2018stacked}&Triplet &46.4&77.4&87.2&34.4&63.7&75.7\\
{\bfseries SCAN (I2T)}&{\bfseries Max Polynomial Loss}&{\bfseries46.9}&{\bfseries 77.7}&{\bfseries87.6}&{\bfseries34.4}&{\bfseries64.2}&{\bfseries75.9}\\
\hline
\end{tabular}}
\end{center}

\caption{Experimental results on MS-COCO.}
\label{t2}
\end{table*}

\subsection{Implementation Details}
{\bfseries Image-Text Matching.} We evaluate our polynomial loss on two standard benchmarks: MS-COCO~\cite{lin2014microsoft} and Flickr30K~\cite{young2014image}; {\bfseries MS-COCO} dataset contains 123,287 images, and each image comes with 5 captions. We mirror the data split setting of \cite{lee2018stacked}. More specifically, we use 113,287 images for training, 5,000 images for validation and 5,000 images for testing. We report results on both 1,000 test images (averaged over 5 folds) and full 5,000 test images. {\bfseries Flickr30K} dataset contains 31,783 images, each image is annotated with 5 sentences. Following the data split of \cite{lee2018stacked}, we use 1,000 images for validation, 1,000 images for testing and the remaining for training.

Our implementation follows the practice in Stacked Cross Attention Network (SCAN)~\cite{lee2018stacked}. SCAN maps image regions and words into a shared embedding space to measure the similarity score between an image and a caption. For fair comparison, we keep the network structure unchanged and replace the loss function with polynomial loss. There are two inputs for SCAN, a set of image features which extracted by a pretrained Faster-RCNN model~\cite{anderson2018bottom} with ResNet-101~\cite{he2016deep}, and a set of word features which encoded by a bi-directional Gated Recurrent Unit (GRU)~\cite{schuster1997bidirectional}. Models are trained from scratch using Adam~\cite{kingma2014adam} with batch size of 128 for both datasets.  For MS-COCO, we start training with learning rate 0.0005 for 10 epochs, and then lower it to 0.00005 for another 10 epochs. For Flickr30K, the learning rate is 0.0002 for 15 epochs, and then lower it to 0.00002 for another 15 epochs. There are two sets of parameters in the polynomial loss, $\{a_p\}$ and $\{b_q\}$. We adopt a heuristic method to select hyper-parameters. Concretely, we first initialize the $G(\cdot)$ to ensure that its curve conforms to the trend in Figure \ref{fig2}. Then, a grid search technology is adopted to select hyper-parameters. We set $P=2$, $\{a_0=0.5, a_1=-0.7,a_2=0.2\}$, $Q=2$ and $\{b_0=0.03, b_1=-0.3,b_2=1.2\}$ for MS-COCO and $P=2$, $\{a_0=0.6, a_1=-0.7,a_2=0.2\}$, $Q=2$, $\{b_0=0.03, b_1=-0.4,b_2=0.9\}$ for Flickr30K. 

{\bfseries Video-Text Matching.} We evaluate our polynomial loss on two popular datasets: ActivityNet-captions~\cite{krishna2017dense} and MSR-VTT~\cite{xu2016msr}. {\bfseries ActivityNet-captions} contains 20K videos, and each video comes with 5 text descriptions. We follow the data split of \cite{Liu2019a}, 10,009 videos for training and 4,917 for testing. {\bfseries MSR-VTT} contains 10K videos and each video is associated with about 20 sentences. We follow the data split of \cite{Liu2019a}, 6,513 videos for training, and 2,990 videos for testing.

\begin{table*}
\begin{center}
\setlength{\tabcolsep}{3mm}{
\begin{tabular}{l|l|ccc|ccc}
\hline
\multirow{2}{*}{Methods} & \multirow{2}{*}{Loss Function} &\multicolumn{3}{c}{Video-to-Text} & \multicolumn{3}{c}{ Text-to-Video }\\
\cline{3-8}
&&R@1&R@5&R@10&R@1 &R@5&R@10\\
\hline
DENSE~\cite{krishna2017dense}& Cross-entropy &18.0 &36.0& 74.0& 14.0& 32.0 &65.0\\
HSE (4SEGS)~\cite{zhang2018cross} & Multi-loss & 18.7 &48.1&- &20.5 &49.3&-\\
CE~\cite{Liu2019a} & Triplet&27.9& 61.6& {\bfseries95.0}& 27.3& 61.1 & 94.4\\
	{\bfseries CE } & {\bfseries Max Polynomial Loss}&{\bfseries 27.9}& {\bfseries 61.9} & 94.1&{\bfseries 28.5}& {\bfseries 62.6}& {\bfseries 94.9}\\
\hline
\end{tabular}}
\end{center}

\caption{Experimental results on ActivityNet-captions.}
\label{t3}
\end{table*}

\begin{table*}
\begin{center}
\setlength{\tabcolsep}{3mm}{
\begin{tabular}{l|l|ccc|ccc}
\hline
\multirow{2}{*}{Methods} & \multirow{2}{*}{Loss Function} &\multicolumn{3}{c}{Video-to-Text} & \multicolumn{3}{c}{ Text-to-Video }\\
\cline{3-8}
&&R@1&R@5&R@10&R@1 &R@5&R@10\\
\hline
Minthum wt al.~\cite{mithun2018learning}& Cross-entropy &12.5&32.1& 42.4& 7.0& 20.9&29.7\\
W2VV~\cite{dong2018predicting}& Multi-loss & 11.8 &28.9 &39.1 &6.1&18.7&27.5\\
Dual encoding~\cite{dong2018dual} &Triplet& 13.0 &30.8&43.3&7.7&22.0&31.8\\
CE~\cite{Liu2019a} & Triplet& 34.4& 64.6 & 77.0&22.5& 52.1& 65.5\\
{\bfseries CE }& {\bfseries Max Polynomial Loss}&{\bfseries 36.2}&{\bfseries 71.5}& {\bfseries 82.2}& {\bfseries 25.0}& {\bfseries 55.4}& {\bfseries 68.2}\\
\hline
\end{tabular}}
\end{center}
\caption{Experimental results on MSR-VTT.}
\label{t4}
\end{table*}

We report results on video-text matching task with Collaborative Experts (CE)~\cite{Liu2019a} framework. CE is a framework that aggregated various pretrained features of a video into a dense representation before mapping to the shared embedding space. We keep the network structure unchanged and replace the loss function with polynomial loss. Models are trained from scratch using Adam~\cite{kingma2014adam} with batch size of 64 for both datasets. The learning rate is set to 0.0004. There are two sets of parameters in the polynomial loss, $\{a_p\}$ and $\{b_q\}$. We set $P=2$, $\{a_0=0.5, a_1=-0.7,a_2=0.2\}$, $Q=2$ and $\{b_0=1, b_1=-0.2,b_2=1.7\}$ for ActivityNet-captions and $P=2$, $\{a_0=0.5, a_1=-0.7,a_2=0.2\}$, $Q=2$, $\{b_0=0.03, b_1=-0.3,b_2=1.8\}$ for MSR-VTT.

\subsection{Image-Text Matching Results}
For image-text matching task, we compare the performance of our method with the several state-of-the-art methods, including: PVSE~\cite{song2019polysemous}, VSE++~\cite{faghrivse++}, SCO~\cite{huang2018learning}, RRF~\cite{liu2017learning}, DAN~\cite{nam2017dual}, GXN~\cite{gu2018look}and SCAN~\cite{lee2018stacked}. Table \ref{t1} and Table \ref{t2} summarize the results of our method on the Flickr30K and MS-COCO datasets, respectively. We also list the loss function used by various methods. From the table, we can make the following observations:
\begin{itemize}
\item From Table \ref{t1}, we find the proposed method outperforms the baseline SCAN at all metrics. Compared with the classical triplet loss, the performance of SCAN with polynomial loss improves R@1 by  3.6\% for text to image retrieval and 1.5\% for image to text retrieval on Flickr30K.
\item Table \ref{t2} summarizes the results on MS-COCO dataset. From Table \ref{t2}, we can observe that the proposed method outperforms the state-of-the-art approaches, especially for R@1.
By replacing the triplet loss with our $L_{Max}$, the performance of SCAN improves 1.9\% on image to text retrieval (R@1) and 2.4\% on text to image retrieval (R@1) on 1K test images.
\item Classical triplet loss tries to sample the informative pairs from redundant pairs, but treats positive and negative pairs equally. In contrast to it, the proposed polynomial loss assigns appropriate weight value to the positive and negative pairs, and the weight value is related to its similarity score. The proposed method can simultaneously select and weight informative pairs. Extensive experimental results demonstrated that the proposed polynomial loss improved the matching performance effectively.
\end{itemize}

\subsection{Video-Text Matching Results}
We evaluate the effectiveness of our method on two standard benchmarks: ActivityNet-captions and MSR-VTT. We report our results and comparison with current state-of-the-art methods for video-to-text and text-to-video retrievals. The results are summarized in the Table~\ref{t3} and Table~\ref{t4} for ActivityNet-captions and MSR-VTT datasets, respectively.

In order to promote a comprehensive comparison, we list existing state-of-the-art results on these datasets, including: DENSE~\cite{krishna2017dense}, HSE~\cite{zhang2018cross}, CE~\cite{Liu2019a} for ActivityNet-captions dataset and Minthum \etal~\cite{mithun2018learning}, W2VV~\cite{dong2018predicting}, CE~\cite{Liu2019a} and Dual encoding~\cite{dong2018dual} for MSR-VTT dataset. Furthermore, we list the loss function used by various methods. From Table \ref{t3} and Table \ref{t4}, we can observe that our method outperforms the baselines on all measures, and achieves the new state-of-the-art performance on the video-text matching task. When compared with CE (Triplet) which uses the same video and sentence encoders with our method, our method improves 2.5\% on text-to-video (R@1) task on the MSR-VTT dataset. Our method outperforms the CE on all metrics on the ActivityNet-captions dataset. The performance gap between CE (Triplet) and CE (Max Polynomial Loss) shows the effectiveness of our polynomial Loss.

\begin{figure}
	\centering
\includegraphics[width=\linewidth]{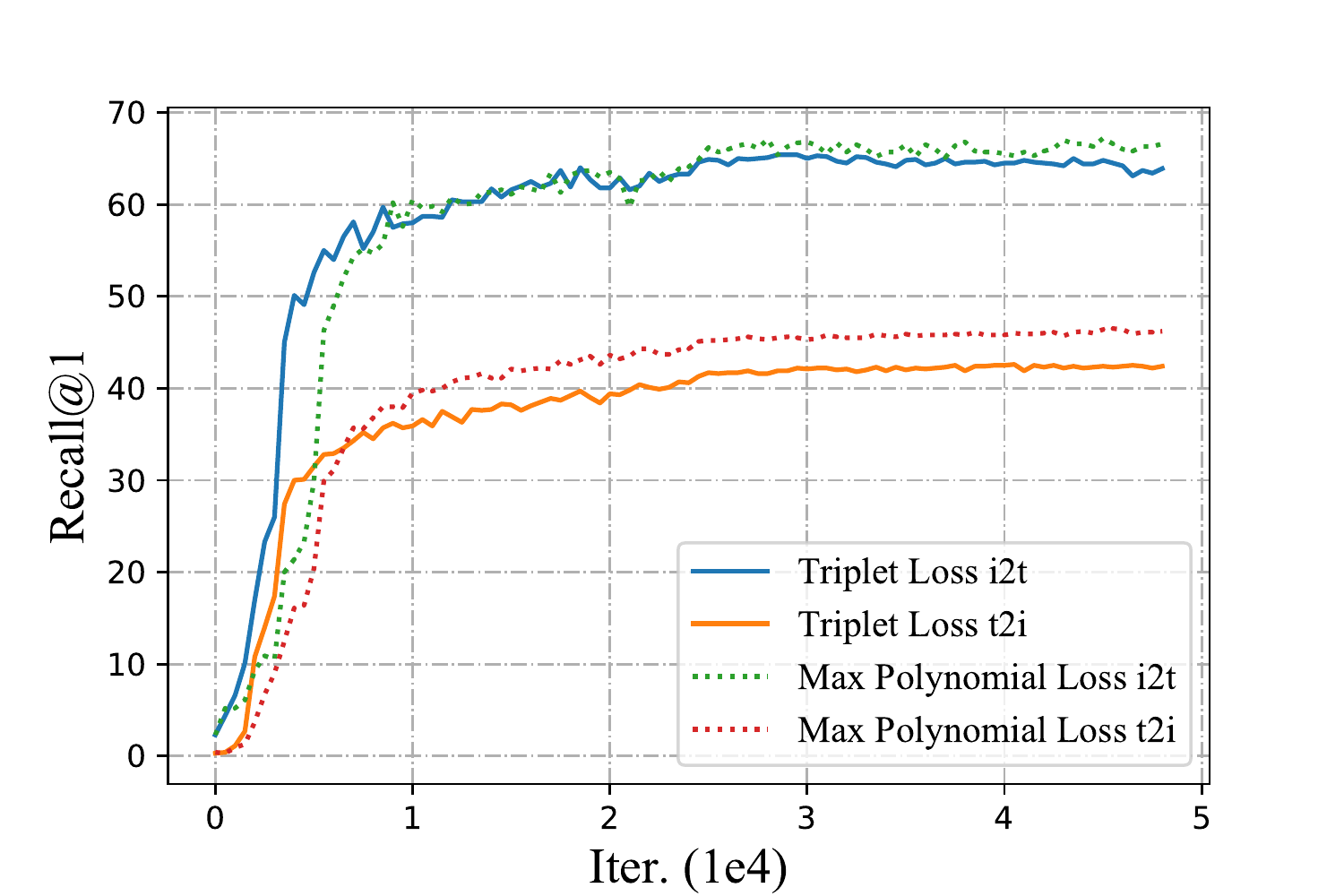}
\caption{Triplet loss vs. Max Polynomial Loss on Flickr30K validation set. By replacing the loss function with our polynomial loss, the performance of SCAN is further improved.}
\label{fig4}
\end{figure}

\begin{figure}
	\centering
\includegraphics[width=\linewidth]{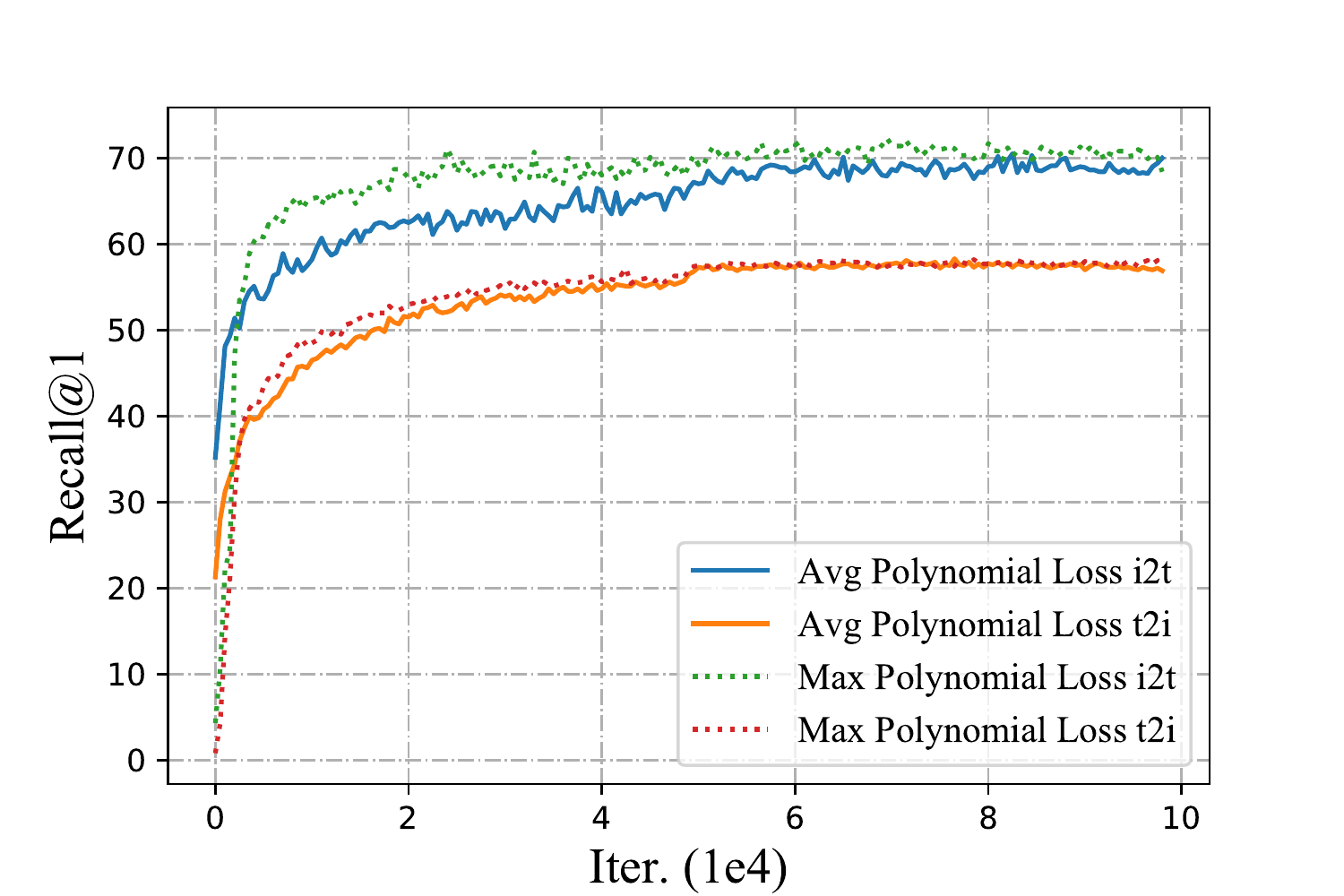}
\caption{Analysis of the behaviors of the Max and Avg polynomial loss on MS-COCO validation set. }
\label{fig5}
\end{figure}

\begin{figure*}
\centering
\includegraphics[width=0.95\linewidth]{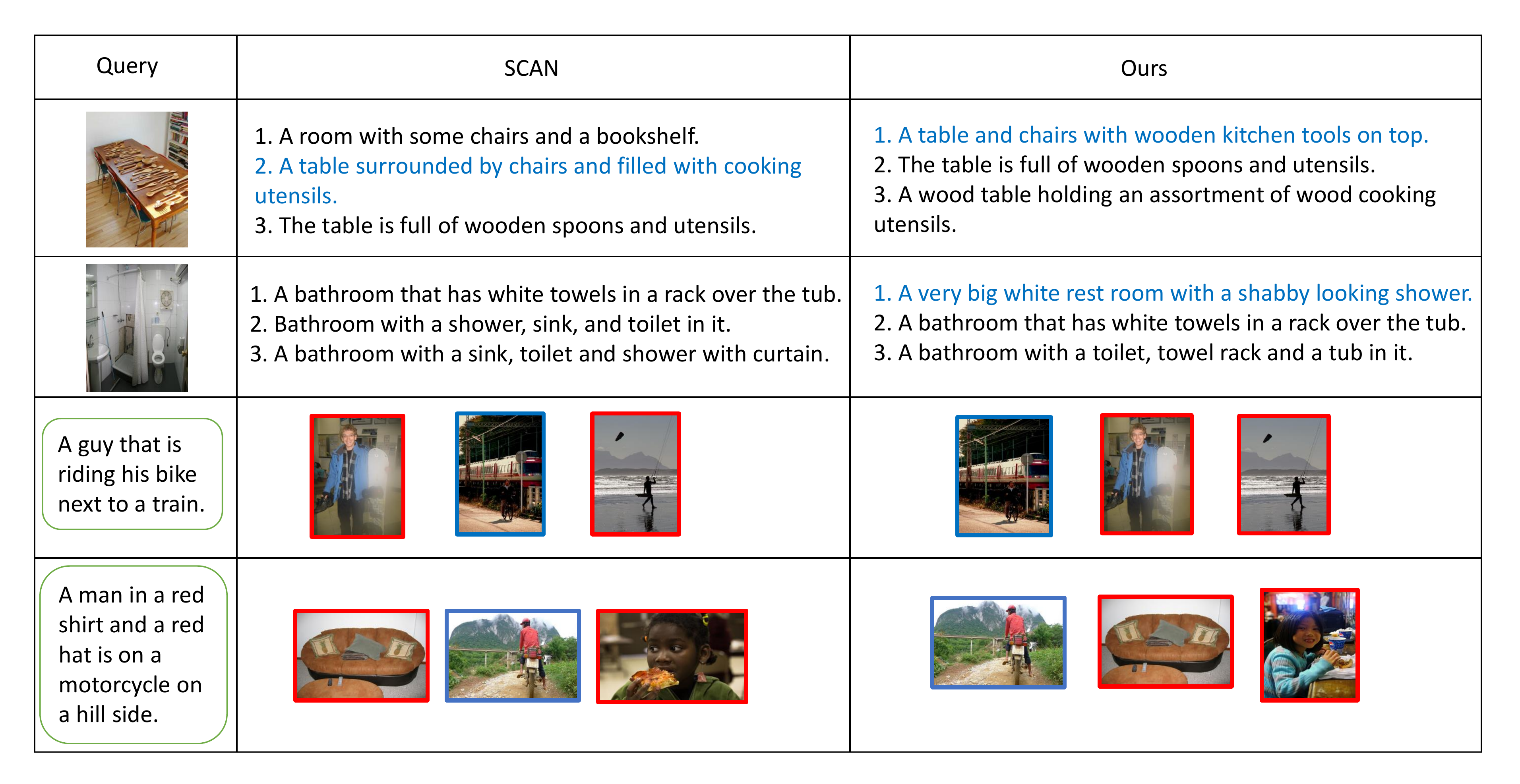}
\caption{Qualitative results on MS-COCO. For each query, we report top-3 ranked results. Predictions ordered by decreasing similarity score, with true matches are shown in blue. For text-to-image retrieval, the true and false matches are outlined in blue and red boxes, respectively.}
\label{fig6}
\end{figure*}

\subsection{Ablation Study}

{\bfseries Parameter Analysis.} There are two sets of parameters in the polynomial loss, $\{a_i\}$ and $\{b_j\}$. It is worth exploring to seek a set of parameters to make the model converge faster and achieves better performance. Since the number of hyper-parameters is too large, it is almost impossible to analyze the sensitivity of the hyper-parameters one by one, so we mainly analyze the sensitivity of several hyper-parameters with great influence. $P$ and $Q$ respectively determines the highest power of $G_{Pos}$ and $G_{Neg}$, which has a direct impact on the number of hyper-parameters. Therefore, we first fixed them to 2. In practice, we find that model performance is most sensitive to parameters $\{b_q\}$, thus we mainly analyze the sensitivity of parameters $\{b_q\}$. We test the effect of $b_1$ and $b_2$ by fixing $b_0=0.03$, results are summarized in Table \ref{t5}. $b_1$ and $b_2$ impact the hard level of negative pairs, the model is sensitive to different values. However, all of these combinations outperform the baseline, which demonstrates the superiority of our approach.


{\bfseries Triplet Loss \emph{vs.} Max Polynomial Loss.} Triplet loss is the most frequently used loss function for cross-modal matching tasks. Its effectiveness has been proved by many works, such as \cite{faghrivse++}. In this section, we further analyze the effectiveness of the proposed polynomial weighting mechanism. Max polynomial loss only includes the hardest negative pairs, which can be considered as a weighted version of triplet loss. We compare the max polynomial loss with triplet loss on MS-COCO dataset, the results are shown in Figure \ref{fig4}. From the results, we find the max polynomial loss converges faster than triplet loss and achieves a better result, which proven the superiority of our polynomial weighting mechanism.


{\bfseries Max \emph{vs.} Avg Ploynomial Loss.} In this section, we further analyze the effectiveness of the proposed Max and Avg polynomial loss. Max polynomial loss weights the positive pair and the hardest negative pair for each anchor, which can be considered as a weighted version of the hardest triplet loss. In contrast, average polynomial loss contains all of informative negative pairs and assigns different weight values for them. Since the max polynomial loss only utilizes a subset of informative pairs so that its computational complexity is lower than the average polynomial loss which contains all of informative negative pairs.

\begin{table}
\begin{center}
\setlength{\tabcolsep}{2mm}{
\begin{tabular}{l|c|cccc}
\hline
Tasks & \diagbox{$b_1$}{$b_2$}& 1.5&1.7&1.8 & 1.9 \\
\hline
\multirow{3}{*}{Video-to-Text}&-0.2&36.0&35.5&36.1&35.5\\
&-0.3& 34.6&35.0&{\bfseries 36.2}&35.6\\
&-0.4&34.0&35.3&35.6&35.3\\

\hline
\multirow{3}{*}{Text-to-Video}&-0.2&25.0&25.0&24.8&25.0\\
&-0.3&24.8& 24.9&{\bfseries 25.0}&24.8\\
&-0.4&24.6&24.8&24.7&24.9\\
\hline
\end{tabular}}
\end{center}

\caption{The effect of $b_1$ and $b_2$ on MSR-VTT dataset.}
\label{t5}
\end{table}
Figure \ref{fig5} shows the performance of two functions on MS-COCO dataset. From the results, we find the average polynomial loss is converges faster than the max polynomial loss at the first few iterations. The reason is that the average polynomial loss contains more informative pairs. However, the final performance of max polynomial loss is slightly better than average polynomial loss. This is possibly due to the unreasonable parameter setting. Since average polynomial loss contains too many negatives pairs, it is difficult to find a set of parameters $P_{Neg}$ to fit all informative negative pairs.

\subsection{Qualitative Results}
In this section, we perform visualizations Top-3 retrieval results for a handful of examples on MS-COCO. Both qualitative results of the image-to-text retrieval and the text-to-image retrieval are shown in Figure \ref{fig6}, which qualitatively illustrate the model behavior. Predictions are ordered by decreasing similarity score, with correct labels are shown in blue. From Figure \ref{fig6}, we can observe that by replacing the loss function with our polynomial loss, the performance of SCAN is further improved. 


\section{Conclusion}
We have developed a universal weighting framework for cross-modal matching, which defines a weight function for the positive and negative pairs respectively. Universal weighting framework provides a powerful tool to analyze the interpretability of various loss functions. Furthermore, we proposed a polynomial loss function under the universal weighting framework, which can effectively sample and weight the informative pairs. Experimental results on four cross-modal matching benchmarks have demonstrated the proposed polynomial loss significantly improve the matching performance. In future work, we would like to investigate the potential of more advanced weighting functions for cross-modal matching.


\noindent\textbf{Acknowledgments.}
This work was supported in part by  the National Key Research and Development Program of China (No. 2018AAA0102200); the National Natural Science Foundation of China (No. 61976049 and No. 61632007); the Fundamental Research Funds for the Central Universities (No. ZYGX2019Z015); the Sichuan Science and Technology Program, China, (No. 2019ZDZX0008 and No. 2018GZDZX0032).
{\small

\bibliographystyle{ieee_fullname}
}

\end{document}